\documentclass{ecai} 

\usepackage{latexsym}
\usepackage{amssymb}
\usepackage{amsmath}
\usepackage{amsthm}
\usepackage{booktabs}
\usepackage{enumitem}
\usepackage{graphicx}
\usepackage{color}
\usepackage{multirow}
\usepackage{colortbl}
\usepackage{xcolor}
\usepackage{float}

\newcommand{\BibTeX}{B\kern-.05em{\sc i\kern-.025em b}\kern-.08em\TeX}

\begin{document}


\begin{frontmatter}


\paperid{0783} 


\title{ DivCon: Divide and Conquer for Complex Numerical and Spatial Reasoning in Text-to-Image Generation}


\author[A]{\fnms{Yuhao}~\snm{Jia}\thanks{Corresponding Author. Email: yuhao.jia@emory.edu.}}
\author[B]{\fnms{Wenhan}~\snm{Tan}}

\address[A]{Individual Researcher}
\address[B]{Individual Researcher}


\begin{abstract}
Diffusion-driven text-to-image (T2I) generation has achieved remarkable advancements in recent years. To further improve T2I models' capability in numerical and spatial reasoning, layout is employed as an intermedium to bridge large language models and layout-based diffusion models. However, these methods often rely on closed-source, large-scale LLMs for layout prediction, limiting accessibility and scalability. They also struggle with generating images from prompts with multiple objects and complicated spatial relationships. To tackle these challenges, we introduce a divide-and-conquer approach which decouples the generation task into multiple subtasks. First, the layout prediction stage is divided into numerical \& spatial reasoning and bounding box visual planning, enabling even lightweight LLMs to achieve layout accuracy comparable to large-scale models. Second, the layout-to-image generation stage is divided into two steps to synthesize objects from easy ones to difficult ones. 
Experiments are conducted on the HRS and NSR-1K benchmarks and our method outperforms previous approaches with notable margins. In addition, visual results and user study demonstrate that our approach significantly improves the perceptual quality, especially when generating multiple objects from complex textural prompts.
\end{abstract}

\end{frontmatter}


\section{Introduction}
Large-scale text-to-image generation (T2I) models demonstrate exceptional zero-shot capacity in efficiently generating small quantities and varieties of objects \cite{stable-diffusion,attend-and-excite,eDiff,Scaling-up-gans,Ramesh2021,Saharia2022}. However, these methods still suffer limited capability in generating images from text prompts containing specified object counts, varying sizes, rich details, and complicated spatial relationships \cite{HRS, attend-and-excite,Feng2023,Chen2023,Xiao2023}. Recently, combining large language models (LLMs) and layout-to-image diffusion models has significantly improved T2I models' capability in numerical and spatial reasoning \cite{layoutgpt, attention-refocusing}. These methods leverage the strong reasoning and visual planning ability of LLMs to predict objects' layouts from the text prompt, which provides additional informative conditions for diffusion models to produce higher-quality images.

\begin{figure}[t]
  \centering
  \includegraphics[width=0.49\textwidth]{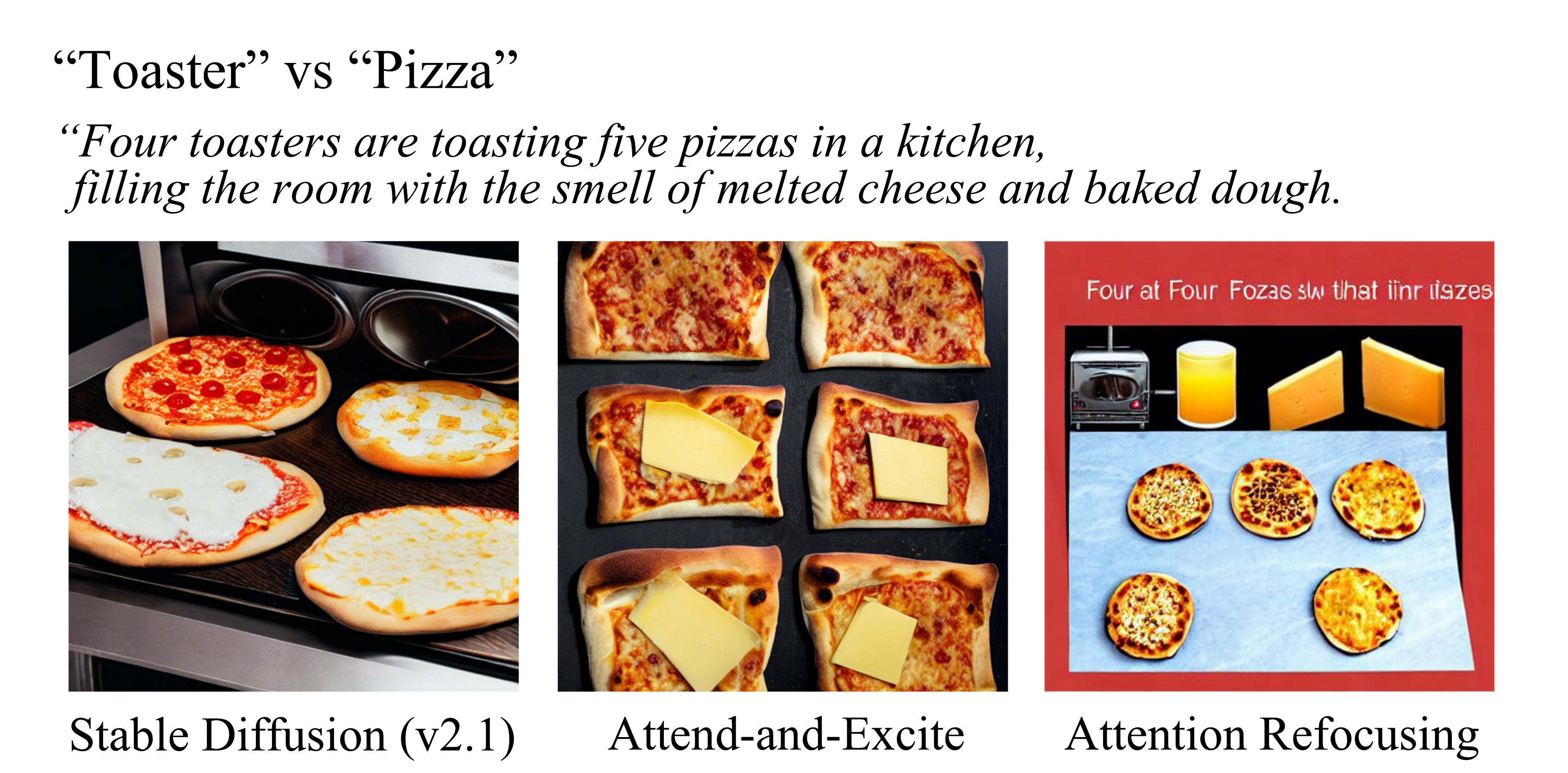}
  \vspace{-0.4cm}
  \caption{A comparison of generation difficulty between ``toaster" and ``pizza".}
  \label{difficulty}
\end{figure}

{Despite recent significant advancements, existing methods still suffer key limitations when dealing with complex prompts involving multiple objects and complex spatial constraints \cite{HRS,attention-refocusing}.  Two major factors hinder their practical effectiveness. \textit{First}, most methods rely on large-scale LLMs such as GPT-4 \cite{openai2024gpt4} to predict layouts\cite{layoutgpt, attention-refocusing, lian2023llmgrounded, qu2023layoutllm}, which, while effective, requires substantial computational resources and limited accessibility. In contrast, whether lightweight, open-source, pretrained language models can perform accurate layout reasoning remains largely unexplored.}
\textit{Second}, layout-to-image diffusion models exhibit varying levels of generative proficiency for objects with different sizes, shapes, and detail complexity. However, most existing methods synthesize all objects simultaneously, without accounting for the varying difficulty of generating different object types. As a result, more complex objects are often poorly synthesized or generated with incorrect counts and spatial positions (\textit{e.g.,} Fig.~\ref{difficulty}), leading to visible inconsistencies in the final images.

\begin{figure}[t]
  \centering
  \includegraphics[width=0.50\textwidth]{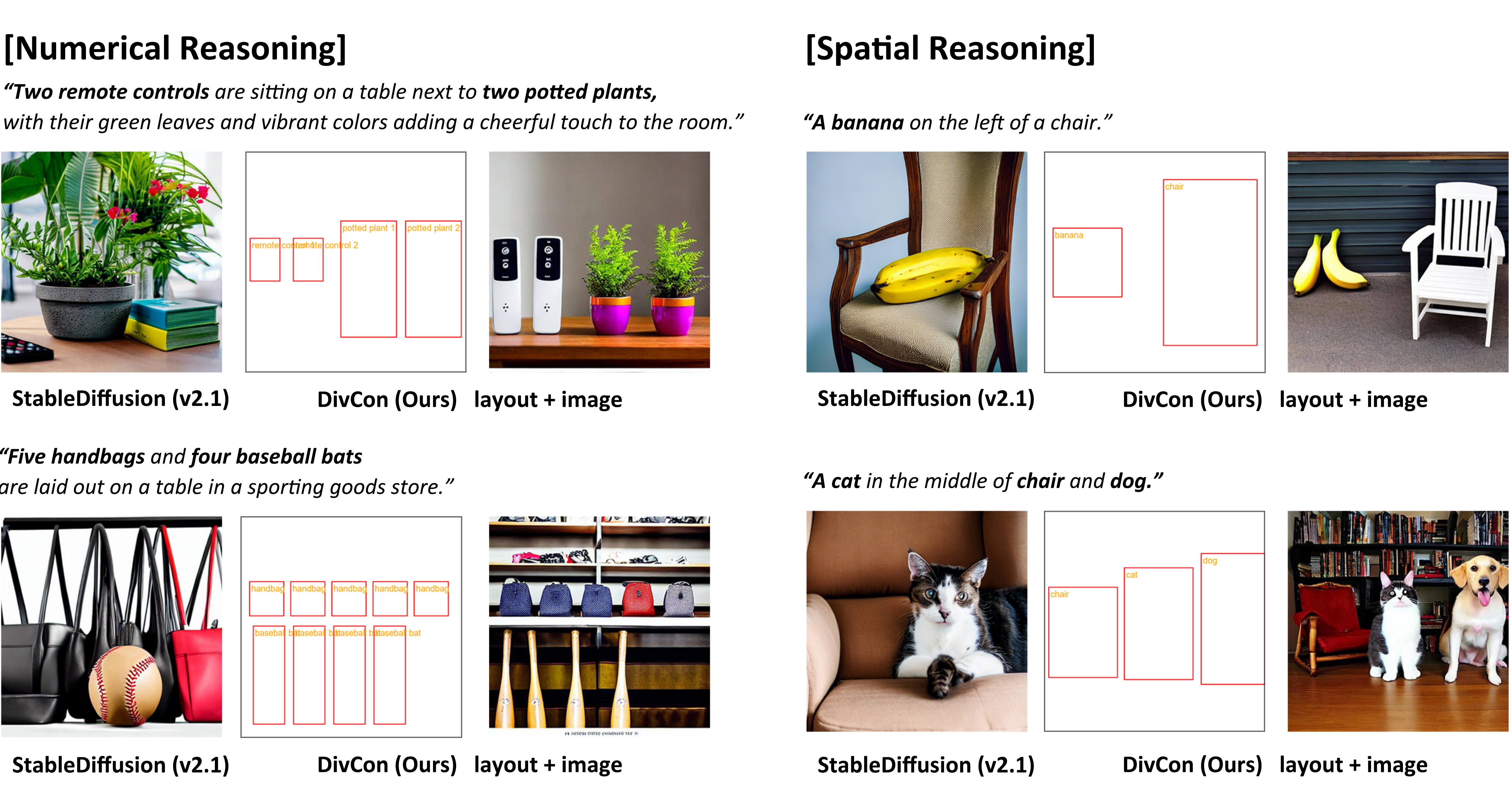}
  \vspace{-0.4cm}
  \caption{Layouts and images generated by our DivCon. DivCon enhances the capability of text-to-image diffusion models to understand complex numerical and spatial relationships in the text.}
  \label{intro_fig}
\end{figure}

To address this issue, we propose \textit{DivCon}, a training-free approach that applies the \textit{divide-and-conquer} strategy to both layout prediction and image generation stages (Fig.~\ref{intro_fig}). 
{In the layout prediction stage, we leverage a lightweight, open-source language model to first perform numerical and spatial reasoning from the text prompt, extracting object counts and positional relations. We then use these reasoning results to guide visual planning through constrained decoding. This decoupled structure significantly improves the numerical and spatial accuracy of the predicted layouts. Despite using only smaller-scale models, our approach achieves layout prediction performance comparable to methods relying on large LLMs, while requiring far fewer computational resources.}
During layout-to-image generation, the diffusion model is first executed to process the layout to generate all objects. Then, the consistency between the resultant objects and the text prompt is calculated. With objects of low consistency being highlighted, the diffusion model is executed for another time to encourage the network to focus more on these difficult objects. By diving layout-to-image generation into two steps, objects with different levels of difficulties can be synthesized progressively for higher quality. Extensive experiments have demonstrated that our method DivCon has significantly improved the quality of generated images and achieves state-of-the-art performance on the HRS\cite{HRS} and NSR-1K\cite{layoutgpt} benchmark datasets.

Our main contributions can be summarized as follows:
\begin{itemize}
    \item We propose DivCon, a divide-and-conquer approach for layout-based text-to-image generation by dividing the complicated task into multiple subtasks.
    {\item To improve layout prediction accuracy with lightweight LLMs to reduce computational cost, we divide layout prediction into numerical \& spatial reasoning and bounding box planning.}
    \item To generate high-fidelity images from the layout, we divide layout-to-image generation into two steps to synthesize objects with different levels of difficulty.
    \item Experimental results show that our approach achieves significant performance gains, surpassing current state-of-the-art models.
\end{itemize}

\begin{figure}[t]
  \centering
  \includegraphics[width=0.46\textwidth]{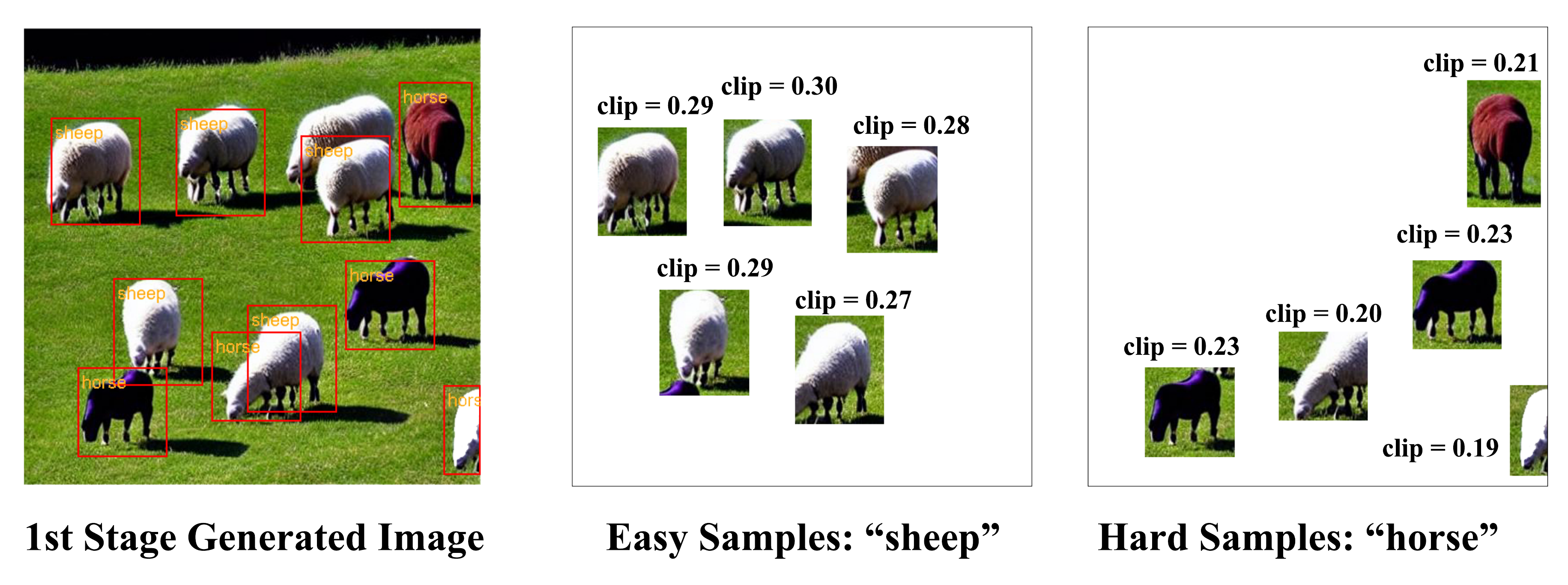}
  \vspace{-0.3cm}
  \caption{Comparison of easy and hard samples in a generated image.}
  \vspace{0.3cm}
  \label{easy/hard samples}
\end{figure}

\begin{figure*}[tb]
  \centering
  \includegraphics[width=1\linewidth]{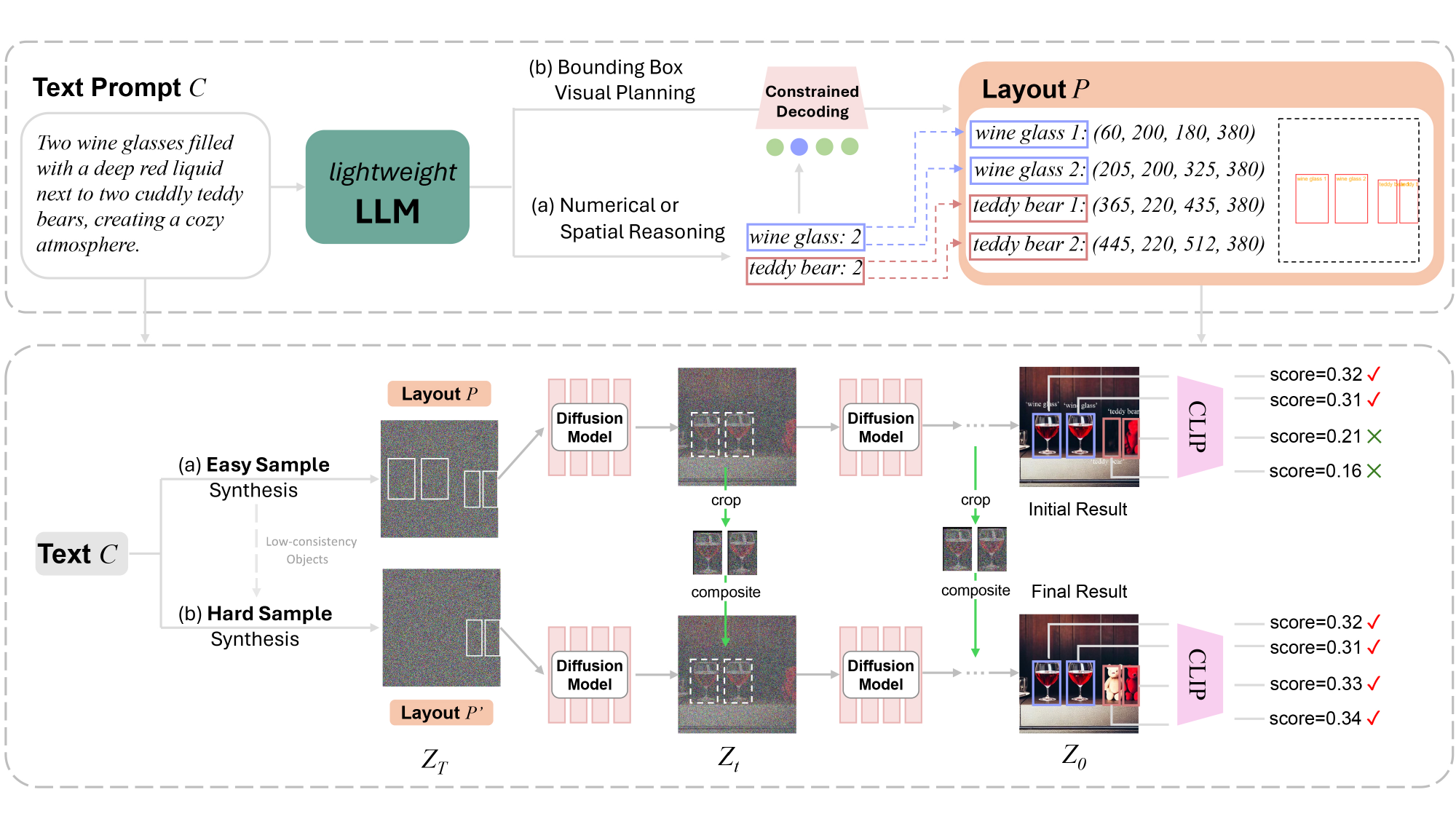}
  \vspace{-0.3cm}
  \caption{\textbf{The proposed DivCon framework.} The layout prediction stage (stage 1) is divided into numerical \& spatial reasoning and bounding box planning. The layout-to-image generation stage (stage 2) is divided into two steps to generate objects with different difficulty levels separately. 
  }
  \label{method}
  \vspace{-0.1cm}
\end{figure*}

\section{Related Work}
In this section, we review three main categories of text-to-image generation methods: text-to-image generation (T2I) models, layout-conditioned T2I models, and text-to-layout-to-image generation models.

\subsection{Text-to-Image (T2I) Generation Models}

The evolution of large-scale T2I synthesis demonstrates a significant leap forward in the realm of generative models, enabling the creation of detailed and coherent images directly from textual descriptions \cite{eDiff, make-a-scene, Ramesh2021, Scaling-up-gans, Saharia2022, yu2022scaling}. Milestone methods in this area are Generative Adversarial Networks (GANs) \cite{Scaling-up-gans, StyleGAN-Ts}, Variational Autoencoders (VAEs) \cite{chang2023muse, ding2022cogview2, Ramesh2021, yu2022scaling}, and diffusion models \cite{eDiff, ho2020denoising, nichol2022glide, ramesh2022hierarchical, Saharia2022}. GANs often struggle with training stability and mode collapse, whereas VAEs may sometimes produce overly smooth or blurry outputs due to their reconstruction loss.  Recently, diffusion models are highly valued for their robustness and ability to generate detailed and realistic images. Particularly, Stable Diffusion \cite{stable-diffusion}, which utilizes text embeddings extracted by CLIP \cite{CLIP} as conditions, has demonstrated remarkable performance for open-world image synthesis. Despite promising results, a text prompt is usually difficult to provide fine-grained control of the generated images.

\subsection{Layout-conditioned T2I Models}

To improve the controllability of T2I models, prompts like layouts are employed as additional conditions for T2I models.
Existing layout-conditioned generation models can be categorized into three branches. One branch of methods train external networks on layout-images pairs \cite{li2023gligen, zhang2023adding, nichol2022glide, avrahami2023spatext, yang2022reco}. For instance, GLIGEN incorporates a trainable gated self-attention layer that transforms bounding boxes into conditional input for Stable Diffusion. Another branch of methods, including attention-refocusing \cite{attention-refocusing}, layout-predictor \cite{layout-predictor}, direct-diffusion\cite{ma2023directed}, layout-guidance\cite{Chen2023} and BoxDiff \cite{xie2023boxdiff}, directly manipulate attention maps to optimize the sampling process of the diffusion model. Since these methods generate all objects through a single forward pass of the diffusion model, objects with spatial relationships usually suffer limited quality. 
To remedy this, the last branch of methods (including Mixture-of-Diffusion\cite{mixture-diffusion}, MultiDiffusion \cite{bar-tal2023multidiffusion}, and LLM-grounded Diffusion \cite{lian2023llmgrounded}) first synthesize each object (\emph{i.e.}, bounding box) separately and then compose them to produce the final result. Nevertheless, the computational cost of these methods increases with the number of objects such that the computational overhead is considerable for a text prompt containing plenty of objects. In addition, the relationship between objects is disrupted and cannot be well maintained in the generated images.

\subsection{Text-to-Layout-to-Image Generation Models}

Current LLMs have demonstrated strong numerical and spatial reasoning capabilities in visual planning and layout generation. Inspired by this, several efforts are made to integrate LLMs to generate layouts from texts, and then utilize layout-conditioned diffusion models for image generation. Attention-refocusing \cite{attention-refocusing} and LLM-grounded Diffusion \cite{lian2023llmgrounded} explore in-context learning of GPT-4 for layout prediction. LayoutGPT \cite{layoutgpt} retrieves similar samples from a database as input examples for LLMs. LLM-layout-generator \cite{LLM-layout-generator} leverages Chain-of-Thought (CoT) prompting \cite{CoT} to improve the layout prediction accuracy. 
{However, these methods all rely on large LLMs such as GPT-4 to directly infer object layouts through end-to-end decoding, which is costly to deploy. To the best of our knowledge, using lightweight LLMs for layout prediction in this setting is still under-investigated.}
{In this paper, we propose a divide-and-conquer approach that can be applied to lightweight LLMs to enhance the spatial and reasoning accuracy for the layout prediction.}
In addition, during layout-to-image generation, the objects are synthesized progressively according to their difficulties, which achieves not only superior fidelity but also significant computational cost savings.

\section{Method}

As illustrated in Fig.~\ref{method}, our framework comprises two stages. Given a text prompt, a lightweight LLM is first adopted to predict a layout. Then, the resultant layout is fed to the diffusion model as an additional condition with the text prompt to achieve text-to-image generation. More specifically, each stage is further divided into two steps to handle subtasks. 

\subsection{Text-to-Layout Prediction}
{Empirically, we observe that while lightweight LLMs possess basic numerical and spatial reasoning capabilities, these models suffer unstable and inaccurate results for layout generation that requires integrated visual planning. To address this, we decouple layout prediction into two explicit stages. First, a small-scale and quantized LLM is prompted to extract object-level information from the text prompt, including either numerical counts or spatial positions. Then, the same model is prompted to perform visual planning for the layout, guided by the reasoning output through constrained decoding.}

\subsubsection{Numerical and Spatial Reasoning.} 
Given a text prompt \(C\) describing an image, a {lightweight} LLM is adopted to predict a set of structured strings \(O\) {containing either numerical or spatial information about objects in the scene}. If the prompt primarily focuses on counting relationships among objects, {the output is parsed into a set of numerical constraints \(O_{\text{num}} = \{(c_j, q_j)\}_{j=1}^{m}\}\), where \(c_{j}\) denotes the object category and \(q_{j}\) is its inferred quantity.}
If the prompt mainly describes spatial relationships, 
{the output is parsed into a set of spatial constraints \(O_{\text{spa}} = \{(o_j, p_j)\}_{j=1}^{n}\), where \(o_{j}\) is the object instance and \(p_{j}\) is its approximate spatial position (e.g., \texttt{"top-left"}). These reasoning results are later parsed and converted into token-level constraints that guide layout generation.}
\subsubsection{Bounding Box Prediction with Constrained Decoding}
After numerical and spatial reasoning, the meta prompt \(C\) {are fed to the LLM again with constrains \(O\)} to predict a list of name-box pairs \(P=\{b_{i} | i= 1, 2, ..., n\}\). Each pair \(b_{i}=\{n_{i}:(x^{(i)}_1, y^{(i)}_1, x^{(i)}_2, y^{(i)}_2) | i= 1, 2, ..., n\}\) consists of the object name \(n_{i}\) and its bounding box's coordinates. 
{To enforce consistency with the predicted constraints, we apply a constrained decoding strategy that inserts structured prefix control tokens during generation. For numerical constraints \(O_{\text{num}} = \{(c_j, q_j)\}_{j=1}^{m}\), we construct a sequence of prefix token sequences:
\begin{equation}
T_{\text{num}} = [\underbrace{\mathtt{c_1{:}}, \dotsc, \mathtt{c_1{:}}}_{q_1\ \text{times}}, \dotsc, \underbrace{\mathtt{c_m{:}}, \dotsc, \mathtt{c_m{:}}}_{q_m\ \text{times}}]    
\end{equation}
where each prefix token sequence represents a control string (e.g., \texttt{"apple:"}) associated with category \(c_j\), and is inserted before decoding each bounding box. This ensures that each category \(c_j\) appears exactly \(q_j\) times in the output.}
{
For spatial constraints \(O_{\text{spa}} = \{(o_j, p_j)\}_{j=1}^{n}\),  we construct a sequence of prefix token sequences:
\begin{equation}
T_{\text{spa}} = [\mathtt{o_1(p_1):}, \mathtt{o_2(p_2):}, \dotsc, \mathtt{o_n(p_n):}]
\label{eq:spatial_prefix}
\end{equation}
where each control prefix string contains both the object name and its position (e.g., \texttt{"apple(top-left):"}). The token sequence of each prefix is inserted before the decoding of the bounding box for each object \(o_j\), guiding the predicted location toward the specific region while enabling the model to freely complete the remaining coordinates. The detailed prompt templates are provided in the Appendix.}

\begin{table*}[t]
\centering
\setlength{\tabcolsep}{4.9pt}
\begin{tabular}{ll
cccc c
cccc c}
\toprule
\multirow{2}{*}{Method} & \multirow{2}{*}{Setting} 
& \multicolumn{4}{c}{\textbf{HRS (numerical)}} & \textbf{HRS (spatial)} 
& \multicolumn{4}{c}{\textbf{NSR-1K (numerical)}} & \textbf{NSR-1K (spatial)} \\
\cmidrule(lr){3-6} \cmidrule(lr){7-7} \cmidrule(lr){8-11} \cmidrule(lr){12-12}
& & Prec. & Rec. & F1. & Acc. & Acc. & Prec. & Rec. & F1. & Acc. & Acc. \\
\midrule
TinyLlama & 1.1B, FP16 & 71.80 & 43.04 & 53.82 & 17.87 & 17.60 & 90.38 & 87.46 & 88.89 & 70.77 & 16.25 \\
\rowcolor{gray!10}
TinyLlama + DivCon & 1.1B, INT4 & 81.56 & 65.14 & 72.43 & 31.11\textcolor{green!60!black}{\fontsize{6.5}{1}\selectfont~(+13.24)} & 34.21\textcolor{green!60!black}{\fontsize{6.5}{1}\selectfont~(+16.61)} & 93.45 & 89.21 & 91.28 & 75.99\textcolor{green!60!black}{\fontsize{6.5}{1}\selectfont~(+5.22)} & 34.60\textcolor{green!60!black}{\fontsize{6.5}{1}\selectfont~(+18.35)} \\\addlinespace[1ex]
Phi-2 & 2.7B, FP16 & 76.36 & 51.43 & 61.46 & 23.08 & 23.56 & 83.35 & 95.63 & 89.07 & 66.52 & 39.58 \\
\rowcolor{gray!10}
Phi-2 + DivCon & 2.7B, INT4 & 81.45 & 53.87 & 64.85 & 34.29\textcolor{green!60!black}{\fontsize{6.5}{1}\selectfont~(+11.21)} & 41.51\textcolor{green!60!black}{\fontsize{6.5}{1}\selectfont~(+17.95)} & 87.61 & 96.05 & 91.63 & 80.10\textcolor{green!60!black}{\fontsize{6.5}{1}\selectfont~(+13.58)} & 54.79\textcolor{green!60!black}{\fontsize{6.5}{1}\selectfont~(+20.19)} \\\addlinespace[1ex]
Mistral-7B & 7B, FP16 & 87.47 & 67.73 & 76.35 & 44.86 & 28.27 & 95.39 & 93.38 & 94.38 & 89.06 & 59.36 \\
\rowcolor{gray!10}
Mistral-7B + DivCon & 7B, INT4 & 93.12 & 87.45 & 90.20 & 69.42\textcolor{green!60!black}{\fontsize{6.5}{1}\selectfont~(+24.56)} & 46.96\textcolor{green!60!black}{\fontsize{6.5}{1}\selectfont~(+18.69)} & 97.36 & 94.50 & 95.90 & 92.03\textcolor{green!60!black}{\fontsize{6.5}{1}\selectfont~(+2.97)} & 69.98\textcolor{green!60!black}{\fontsize{6.5}{1}\selectfont~(+10.62)} \\\addlinespace[1ex]
Qwen3-4B & 4B, FP16 & 76.56 & 67.21 & 71.61 & 45.15 & 24.43 & 89.21 & 84.07 & 86.56 & 76.13 & 49.66 \\
\rowcolor{gray!20}
Qwen3-4B + DivCon & 4B, FP16 & 88.94 & 80.95 & 84.76 & 65.21\textcolor{green!60!black}{\fontsize{6.5}{1}\selectfont~(+20.06)} & 49.23\textcolor{green!60!black}{\fontsize{6.5}{1}\selectfont~(+24.80)} & 92.21 & 89.07 & 91.09 & 80.21\textcolor{green!60!black}{\fontsize{6.5}{1}\selectfont~(+4.08)} & 59.34\textcolor{green!60!black}{\fontsize{6.5}{1}\selectfont~(+9.68)} \\\addlinespace[1ex]
Qwen3-8B & 8B, INT4 & 73.27 & 71.16 & 72.20 & 61.16 & 59.39 & 91.86 & 92.04 & 91.95 & 86.89 & 84.45 \\
\rowcolor{gray!20}
\textbf{Qwen3-8B + DivCon} & 8B, INT4 & \textbf{93.28} & 87.76 & \textbf{90.44} & \textbf{74.77}\textcolor{green!60!black}{\fontsize{6.5}{1}\selectfont~(+13.61)} & 81.46\textcolor{green!60!black}{\fontsize{6.5}{1}\selectfont~(+22.07)} & 97.45 & 96.87 & 97.16 & 90.07\textcolor{green!60!black}{\fontsize{6.5}{1}\selectfont~(+3.18)} & \textbf{94.97}\textcolor{green!60!black}{\fontsize{6.5}{1}\selectfont~(+10.52)} \\\addlinespace[1ex]
GPT-4 (reference) & >70B est. & 88.13 & \textbf{90.91} & 89.50 & 74.05 & \textbf{86.07} & \textbf{97.55} & \textbf{98.41} & \textbf{97.98} & \textbf{93.41} & 94.68 \\
\bottomrule
\end{tabular}
\caption{\textbf{Quantitative results comparison} of our DivCon framework with baselines on HRS and NSR-1K benchmarks in layout prediction.}
\label{tab:layout_comparison}
\end{table*}

\subsection{Layout-to-Image Generation}
After layout prediction, the resultant layout is employed as an additional condition to generate the image. As objects in the layout have different levels of difficulty to synthesize, a divide-and-conquer approach is adopted to reconstruct objects from easy ones to hard ones in a progressive manner, as shown in Fig. \ref{method}(b). First, the layout is fed to the diffusion model to synthesize all objects. Then, the consistency between the synthetic objects and the text prompt is calculated. Objects with low consistency are considered as low-fidelity ones and fed to the diffusion model for a second time while maintaining the results of objects with high consistency. With low-fidelity objects being highlighted, the diffusion model is encouraged to focus on these difficult objects in the second forward pass to achieve higher quality.

\subsubsection{First-Round Generation.}
We employ a layout-to-image diffusion model with sampling optimization \cite{attention-refocusing} as the generative model to generate an image conditioned on the text prompt and the predicted layout.
During the denoising process, random noise \(z_{T}\) is fed to the diffusion model to gradually denoise it to an image \(z_{0}\) after $T$ {time steps}.

\subsubsection{Consistency Evaluation.}
In the generated image \(z_{0}\), each object is cropped using the bounding boxes in the layout and denoted as  \(x_{i} \in \mathbb{R}^{4 \times h \times w}\), where \(h \times w\) represents the height and width of the bounding box. Then, the CLIP similarity \(S_{x_{i},o_{i}} \in (-1,1)\) is employed to evaluate the consistency between the cropped object and its name \(o_{i}\) in the text prompt. The higher the similarity is, the higher the fidelity of the corresponding object is. Consequently, the objects with similarities higher than a specified threshold can be considered as easy samples that are "good enough". In contrast, other objects with low similarities are considered as hard samples that are more difficult to synthesize (\textit{e.g.} "sheep" and "horse" in Fig. \ref{easy/hard samples}). Next, a binary mask \(M\) is produced with bounding boxes of easy objects set to 1 and the rest regions set to 0. 

\subsubsection{Second-Round Refinement.}
The objective of second-round refinement is to preserve the “good enough" objects while reproducing the remaining ones to ease the difficulty of the diffusion model. To this end, a straightforward approach is to retrieve the latent representation of the objects in the first-round denoising process. {Nevertheless, directly freezing the object regions during the denoising process produces inharmonious artifacts around the boundaries.}
{To address this issue, we retrieve the latent representation \(z_{T}\) of all time steps during the denoising process of the first-round generation. }
Then, the binary mask \(M\) is adopted to crop high-fidelity regions in the latent representation. At each time step \(T\) of the second-round forward pass, these cropped representations are employed to replace the corresponding regions in the noises:
\begin{equation}
z'_{T} = z_{T} \circ M + z'_{T} \circ (1-M)
\end{equation}
where \(\circ\) is element-wise multiplication and \(z'_{T}\) is the noise at \(T\) step of the second-round forward pass. With the latent representation of high-fidelity objects (masked by $M$) being retrieved, these objects can be well preserved to encourage the model to focus on reconstructing the remaining ones.
\textcolor{black}{Meanwhile, the bounding boxes of ``bad samples" \(P'\) are passed to the diffusion model as the layout conditions, enabling the model to concentrate more on refining \(P'\) to further improve the overall quality of the generated image. A more comprehensive analysis of the underlying mechanism is provided in the Appendix.}

\begin{table*}[t]
\centering
\setlength{\tabcolsep}{5.5pt}
\begin{tabular}{p{4.3cm}ccccc ccccc}
\toprule
\multirow{2}{*}{Method} 
& \multicolumn{4}{c}{\textbf{HRS (Numerical)}} 
& \textbf{HRS (Spatial)} 
& \multicolumn{4}{c}{\textbf{NSR-1K (Numerical)}} 
& \textbf{NSR-1K (Spatial)} \\
\cmidrule(lr){2-5} \cmidrule(lr){6-6} \cmidrule(lr){7-10} \cmidrule(lr){11-11}
& Prec. & Rec. & F1 & Acc. & Acc. & Prec. & Rec. & F1 & Acc. & Acc. \\
\midrule
Stable Diffusion 2.1 & 73.97 & 53.93 & 62.38 & 16.20 & 7.96 & 78.94 & 75.42 & 77.41 & 33.99 & 22.97 \\
Attend-and-Excite  & 76.57 & 55.94 & 64.65 & 18.37 & 9.23 & 79.21 & 78.62 & 78.91 & 39.51 & 29.61 \\
Divide-and-Bind & 77.95 & 57.50 & 66.18 & 20.15 & 13.15 & 84.99 & 79.13 & 81.96 & 32.35 & 24.00 \\
Our layout + Layout Guidance & 71.09 & 45.98 & 55.84 & 16.58 & 29.85 & 81.97 & 75.41 & 78.55 & 43.13 & 52.88 \\
Our layout + GLIGEN & 75.34 & 62.14 & 69.01 & 23.72 & 42.16 & 85.13 & 84.97 & 85.05 & 52.21 & 59.72 \\
LayoutLLM-T2I & 74.38 & 61.07 & 67.07 & 26.56 & 47.01 & 83.31 & 84.01 & 83.66 & 55.49 & 67.99 \\
Attention-Refocusing & 77.43 & 61.78 & 68.72 & 25.60 & 48.29 & 84.61 & 85.64 & 85.12 & 56.10 & 69.28 \\
\rowcolor{gray!20}
\textbf{DivCon (Ours)} & \textbf{78.65} &\textbf{ 63.41} & \textbf{70.22} & \textbf{29.97} & \textbf{53.96} & \textbf{85.41} & \textbf{87.06} & \textbf{86.22} & \textbf{57.35} & \textbf{71.65} \\
\bottomrule
\end{tabular}
\caption{\textbf{Quantitative results comparison} of our method with baselines on HRS and NSR-1K benchmarks in image generation.}
\label{tab:image_comparison}
\end{table*}

\section{Experiments}
In this section, we provide an evaluation of our method and compare it with state-of-the-art T2I models. Ablation experiments are then conducted to demonstrate the effectiveness of our divide-and-conquer approach in layout prediction and layout-to-image generation stages.
\subsection{Experiment Setup}

\subsubsection{Benchmarks and Baselines}

To evaluate the performance of spatial and numerical reasoning in {both layouts and images} generation tasks, we utilize benchmarks \textbf{HRS}~\cite{HRS} and \textbf{NSR-1K}~\cite{layoutgpt}. The HRS dataset contains 3,000 prompts (labeled with objects' names and specific counts) for numerical relationships and 1,002 prompts (labeled with objects' names and corresponding positions) for spatial relationships. The NSR-1K dataset consists of 762 prompts for numerical relationships and 283 prompts for spatial relationships, all labeled with names, counts, bounding boxes, and ground truth images. 
Compared to NSR-1K, HRS exclusively employs natural language prompts, alongside a larger quantity of objects per sample and more intricate spatial relations. 
{To evaluate layout prediction, we employ several open-source LLMs of varying scales and architectures as layout generators to compare their performance, including TinyLlama \cite{zhang2024tinyllama}, Phi-2 \cite{javaheripi2023phi}, Mistral-7B \cite{jiang2023mistral7b}, Qwen3-4B and Qwen3-8B \cite{qwen3}. 
 We evaluate both the standalone layout prediction performance of these models and their corresponding quantized versions enhanced by the proposed \textit{DivCon} framework. GPT-4 \cite{openai2024gpt4} is included as a reference representing the strongest known performance for LLM's layout prediction. Among them, DivCon with the quantized Qwen3-8B is used to provide layout inputs for the next stage.}
{For image generation,} Stable Diffusion \cite{stable-diffusion}, Attend-and-Excite~\cite{attend-and-excite}, {Divide-and-Bind \cite{li2023divide}, GLIGEN \cite{li2023gligen}, Layout-Guidance \cite{Chen2023}}, LayoutLLM-T2I \cite{qu2023layoutllm}, and Attention-Refocusing \cite{attention-refocusing} are included for comparison. 
{In addition, we also include the dataset in \cite{lian2023llmgrounded} for evaluation, which includes four tasks: negation, generative numeracy, attribute binding, and spatial reasoning. As our method focuses on numerical and spatial reasoning, we include all prompts from these two tasks (100 per task) to construct a benchmark termed LLM-Grounded.}

\subsubsection{Evaluation Metrics}
\textbf{Fidelity.}
To evaluate the fidelity of both the generated {layouts and} images, grounding accuracy between layouts/images and the text prompt is calculated. YOLOv8 \cite{Jocher_Ultralytics_YOLO_2023} is employed to obtain the bounding boxes of the objects in the generated images. For counting, the number of bounding boxes is compared to the ground truths to calculate the precision, recall, F1 score, and accuracy \cite{el-nouby2019tell, fu2020sscr}. For spatial relationships, the relative position between two bounding boxes is adopted following PaintSkills~\cite{cho2022dalleval} and LayoutGPT \cite{layoutgpt}. Specifically, if the horizontal distance between two bounding boxes exceeds the vertical distance, they are denoted as a left-right relationship. Otherwise, they are denoted as a top-bottom relationship. 

\indent\textbf{Quality.}
To assess the quality of generated images, we use FID as the metric~\cite{heusel2017gans}. Lower FID represents higher similarity between generated images and ground truths. 

{\textbf{Efficiency.}
To validate the efficiency and scalability of our method, inference runtime is evaluated on an AWS g5.xlarge GPU instance.}

\begin{figure*}[ht]
  \centering
  \includegraphics[width=1\linewidth]{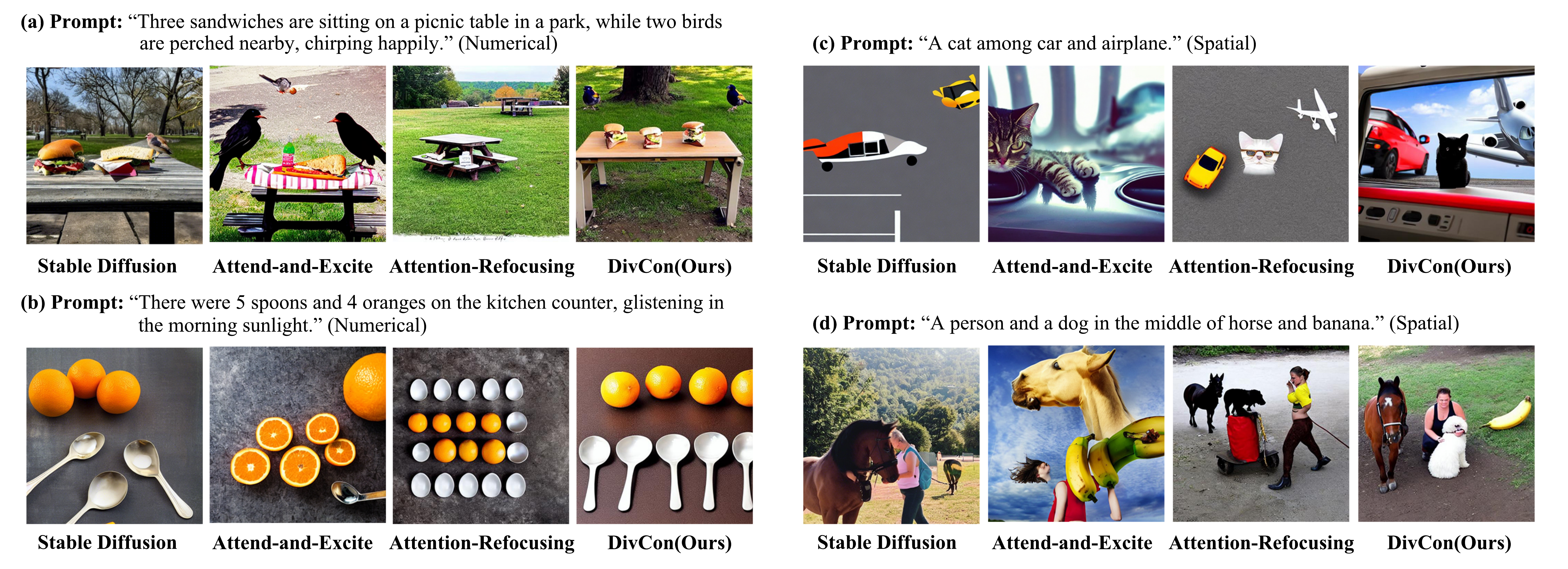}
  \vspace{-0.4cm}
  \caption{Qualitative comparison of numerical \& spatial reasoning between Stable Diffusion, Attend-and-excite, Attention -Refocusing and our DivCon. }
  \label{qualitative1}
\end{figure*}

\subsection{Evaluation Results}
\subsubsection{Quantitative Results for Layout Prediction}
{
We present the results of layout prediction on the HRS and NSR-1K benchmarks in Tab.~\ref{tab:layout_comparison}.
As we can see, all quantized lightweight models, when equipped with our DivCon framework, exhibit substantial improvements across all layout prediction metrics over their original counterparts. For instance, \textbf{Mistral-7B + DivCon} achieves a 24.56\% gain in numerical accuracy on HRS, while \textbf{Qwen3-8B + DivCon} improves spatial accuracy on HRS by 22.07\%. This demonstrates the effectiveness of our divide-and-conquer strategy in enhancing reasoning capabilities of lightweight LLMs.}

{
Notably, our best-performing model, \textbf{Qwen3-8B + DivCon}, surpasses GPT-4 in most HRS’s numerical reasoning metrics, including a 0.94\% gain in F1 score and a 0.72\% gain in accuracy. For HRS's spatial reasoning, while slightly inferior to GPT-4, it achieves comparable accuracy. On NSR-1K, our method matches GPT-4 in both numerical and spatial reasoning. These results demonstrate that lightweight open-source models are capable of matching the performance of much larger LLM models.
}

\subsubsection{Quantitative Results for Image Generation}

We present the results on the HRS and NSR-1K benchmarks in Tab.~\ref{tab:image_comparison}.
{Compared to single-stage text-to-image models, DivCon achieves substantial gains in both numerical and spatial reasoning. On HRS, it improves numerical F1 by over 8\%, numerical accuracy by 9\%, and spatial accuracy by more than 40\%. On NSR-1K, the corresponding improvements reach 8\%, 25\%, and 50\%, respectively. }

{In addition, DivCon also outperforms recent state-of-the-art text-to-layout-to-image models with notable margins. On HRS, it yields a 1.5\% improvement in numerical F1 and a 3.41\% gain in numerical accuracy, while achieving a 5.67\% higher spatial accuracy. On NSR-1K, DivCon maintains strong performance, surpassing prior methods by 1.1\% in numerical F1, 1.25\% in accuracy, and 2.37\% in spatial accuracy. These consistent improvements highlight the effectiveness of our divide-and-conquer strategy in enhancing the fidelity and alignment of generated images under complex reasoning tasks.}

In addition to the HRS and NSR-1K benchmarks, we compare DivCon with LLM-Grounded on numerical and spatial reasoning tasks, as shown in Tab.~\ref{tab:llm-grounded_compare}. While both methods achieve comparable performance in numerical reasoning, DivCon demonstrates a notable 5\% improvement in spatial accuracy, further highlighting its ability to handle spatial relationships in complex text prompts.

\begin{table}[ht]
    \centering
    \setlength{\tabcolsep}{3pt}
    \renewcommand{\arraystretch}{1.3}
    \begin{tabular}{>{\centering\arraybackslash}p{0.23\textwidth} >{\centering\arraybackslash}p{0.1\textwidth} >{\centering\arraybackslash}p{0.06\textwidth}} \hline
         \textbf{Methods}& \textbf{ Numerical}& \textbf{Spatial}\\ \hline
         LLM-Grounded (GPT4) &  84\%& 82\%\\
         \rowcolor{gray!20}
         \textbf{DivCon (Ours)}&  84\%& 87\%\\ \hline
    \end{tabular}
    \caption{Comparison with LLM-Grounded}
    \label{tab:llm-grounded_compare}
\end{table}

\textbf{Image quality} results are shown in Tab. \ref{tab:fid}. As we can see, our DivCon produces comparable FID scores to Stable Diffusion and outperforms attention-refocusing with notable margins. This further demonstrates that our approach is able to produce images with better visual quality while maintaining high prompt fidelity.

\begin{table}[ht]
\small
\setlength{\tabcolsep}{2.5pt}
\centering
\begin{tabular}{>{\centering\arraybackslash}p{0.6\linewidth}>{\centering\arraybackslash}p{0.3\linewidth}}
\toprule
{\bf Methods}& {\bf FID} ($\downarrow$)\\
\midrule
 Stable Diffusion 2.1& 20.77\\
Attention-Refocusing& 21.05\\
\rowcolor{gray!20} DivCon (Ours)& 21.01 \\
\bottomrule
\end{tabular}  
\caption{FID results achieved on NSR-1K.}
\label{tab:fid}
\end{table}

\textbf{Runtime comparisons} of the diffusion process are presented in Tab. \ref{tab:runtime}. Among techniques that require multiple forward passes to synthesize different regions separately, our DivCon produces notable efficiency gains. Particularly, LLM-Grounded takes 134.98s to generate an image, while our method achieves a $1.87\times$ speedup. Compared with layout-conditioned methods with a single forward pass, our method (1 Iter) produces competitive results with comparable efficiency.

To conclude, by dividing the T2I task into simpler subtasks and progressively addressing numerical and spatial reasoning, DivCon demonstrates superior fidelity in interpreting text prompts and generating corresponding layouts and images. 
{Notably, our framework enables lightweight LLMs to achieve layout prediction performance comparable to large-scale models like GPT-4, significantly reducing computational cost without sacrificing accuracy.} 
Moreover, DivCon maintains high visual quality in the generated images while achieving better alignment with the textual descriptions, as reflected in improved spatial and numerical metrics. This highlights the robustness and scalability of our approach in tackling challenging T2I scenarios.

\begin{table}[ht]
\centering
\setlength{\tabcolsep}{3pt}
\small
\begin{tabular}{>{\centering\arraybackslash}p{0.4\linewidth}>{\centering\arraybackslash}p{0.3\linewidth}>{\centering\arraybackslash}p{0.15\linewidth}}
    \toprule
    \textbf{Methods} & \textbf{\#Forward Pass} & \textbf{Runtime} \\
    \midrule
    GLIGEN & 1 & \textbf{33.75 s} \\
    Attention-Refocusing & 1 & 35.71 s \\
    \rowcolor{gray!20} \textbf{DivCon-1 iter (Ours)} & 1 & 37.35 s \\
    \midrule
    Mixture-of-Diffusion  & 3 & 126.41 s \\
    LLM-Grounded  & 3 & 134.98 s \\
    Multi-Diffusion & 3 & 121.68 s \\
    \rowcolor{gray!20} \textbf{DivCon-2 iter (Ours)} & 2 & \textbf{72.01 s} \\
    \bottomrule
\end{tabular}
\caption{Comparison of Inference Runtime.}
\label{tab:runtime}
\vspace{-0.3cm}
\end{table}

\subsubsection{Qualitative Results}
Figure~\ref{qualitative1} illustrates the qualitative comparison of DivCon and baselines. In all cases, our DivCon can accurately predict the object quantities and the spatial relations from the text prompts. 

\textbf{Large Quantity of Objects.}
Given a prompt containing more than one object category (Fig. \ref{qualitative1}), Stable Diffusion \cite{stable-diffusion} and Attend-and-Excite \cite{attend-and-excite} often generate an insufficient quantity of objects, while Attention-Refocusing \cite{attention-refocusing} sometimes omits one category of objects or generates an excessive number of objects. {For instance, the prompt in Fig. \ref{qualitative1}(b) contains five spoons and four oranges. However, Stable Diffusion and Attend-and-Excite cannot synthesize sufficient spoons or oranges. Meanwhile, Attention-Refocusing generates an excess number of objects, with spoons appearing incomplete due to missing parts.} As compared to these methods, our DivCon achieves superior numerical accuracy.

\textbf{Complicated Spatial Relation.}
{Given a prompt containing complex spatial relationships among more than two object categories, previous T2I models struggle to generate correct spatial relations. For example, Stable Diffusion and Attend-and-Excite miss certain object categories (Fig.~\ref{qualitative1}(d)). Meanwhile, Attention-Refocusing tends to produce distortion or blending artifacts around the boundaries of objects (Fig.~\ref{qualitative1}(c)(d)) or merges the color of the banana with the person (Fig.~\ref{qualitative1}(d)). In contrast, our DivCon produces accurate object categories and spatial relationships, exhibiting superior performance in these challenging cases. }

\begin{figure}[th]
  \centering
  \includegraphics[width=0.99\linewidth]{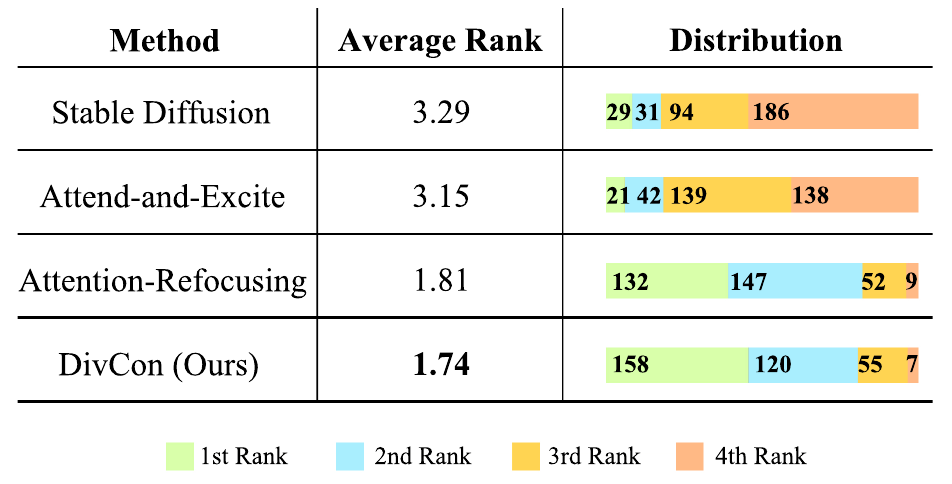}
  \caption{User Study: Comparison of image quality.}
  \label{fig:user study}
\end{figure}

\textbf{User Study.}
We conduct a user study to evaluate the images generated by different methods. To be specific, we recruit 17 people with different backgrounds and asked them to rank the images generated by different methods according to their fidelity to the original prompt. Lower ranks indicates better alignment. A total of 20 examples are randomly selected from the NSR-1K benchmark.
As shown in Fig.~\ref{fig:user study}, our method achieves the best performance in terms of average rank, which demonstrates the superior perceptual quality of our results.

\subsection{Ablation Study}

In this sub-section, we conduct ablation experiments to evaluate the effectiveness of our divide-and-conquer approach in both layout prediction and image generation. Results are presented in Tab. \ref{tab:ablation}. QWen3-8B \cite{qwen3} and image generator in Attention-Refocusing \cite{attention-refocusing} are used as the baseline for two stages. 

\begin{figure}[th]
  \centering
  \includegraphics[width=1\linewidth]{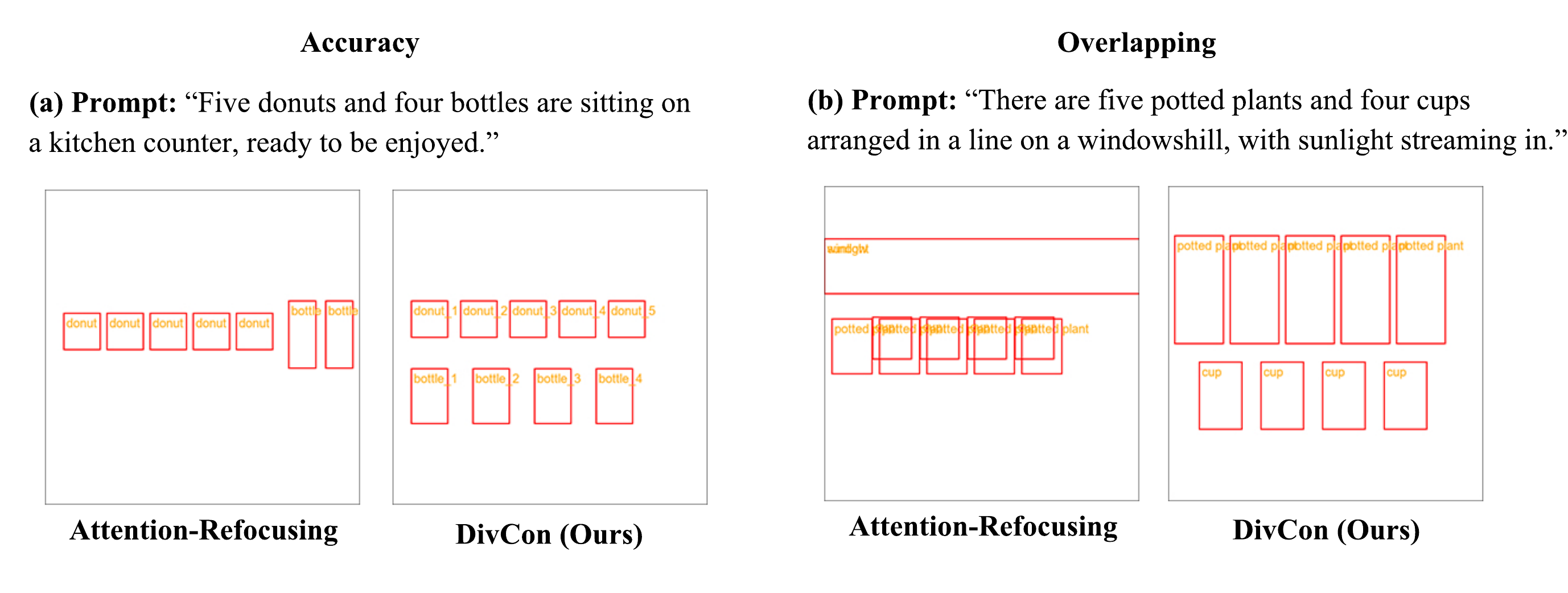}
  \vspace{-15pt}
  \caption{Comparison of the predicted layouts. }
  \label{layout compare}
  \vspace{-0.1cm}
\end{figure}

\textbf{Divide-and-Conquer in Layout Prediction.}
To demonstrate the effectiveness of our divide-and-conquer approach in the layout prediction stage, we develop a variant by conducting numerical \& spatial reasoning and bounding box prediction in a single step. As we can see, our divide-and-conquer approach produces consistent improvements across all metrics, with an increase of 7.59\% in numerical F1 score, 8.75\% in numerical accuracy and 12.84\% in spatial accuracy. We further compare the layouts predicted by our DivCon and Attention-Refocusing in Fig. \ref{layout compare}. Compared to Attention-Refocusing, layouts generated by our DivCon not only exhibit higher reasoning accuracy (Fig. \ref{layout compare} (a) but also are more organized with less overlap (Fig. \ref{layout compare} (b)).

\begin{figure}[t]
  \centering
  \includegraphics[width=1\linewidth]{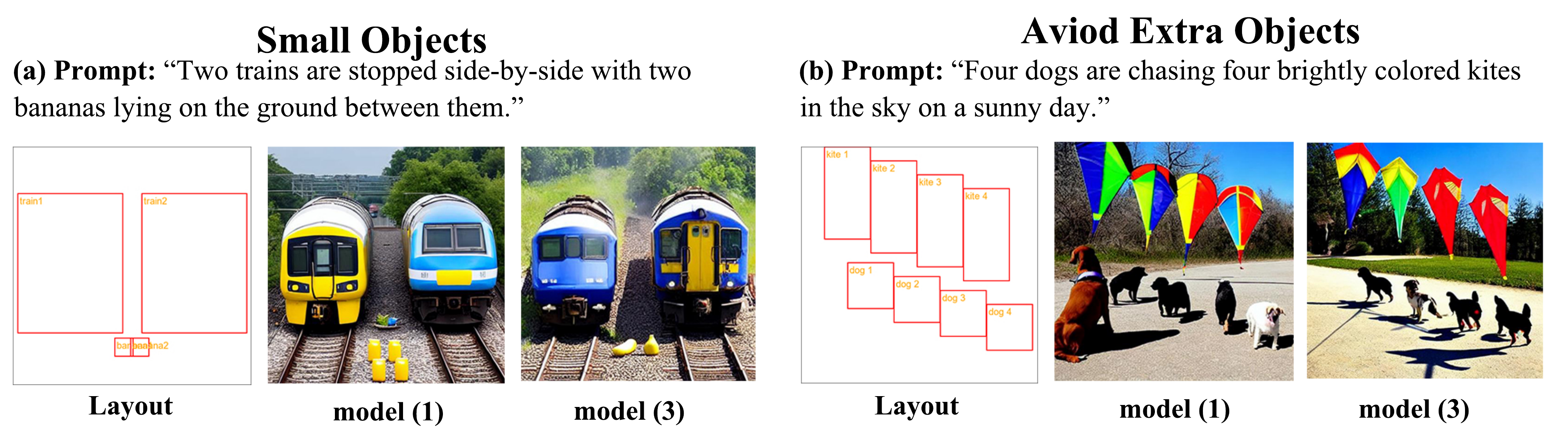}
  \vspace{-0.3cm}
  \caption{Ablation study of our image generation. }
  \label{ablate generation}
  \vspace{0.3cm}
\end{figure}

\textbf{Divide-and-Conquer in Image Generation.} 
To study the effectiveness of our divide-and-conquer approach in the image generation stage, we develop a variant to synthesize the image in a single forward pass. It can be observed from Tab.~\ref{tab:ablation} that the divide-and-conquer approach facilitates our method to notable performance gains in terms of all metrics. We further compare the results produced by our method using prompts containing multiple objects in Fig.~\ref{ablate generation}. As we can see, the divide-and-conquer approach facilitates our method to better reconstruct the objects. 
For example, as shown in Fig.~\ref{ablate generation}, the progressive synthesis enabled by the divide-and-conquer strategy allows DivCon to reconstruct objects more accurately, maintaining both numerical consistency and spatial relationships specified in the prompts. This demonstrates that the divide-and-conquer framework facilitates better alignment and fidelity in text-to-image tasks.

\begin{table}[t]
  \centering
  {\scriptsize
  \begin{tabular}{>{\centering\arraybackslash}p{0.01\linewidth}@{ }>{\centering\arraybackslash}p{0.15\linewidth}@{ }>{\centering\arraybackslash}p{0.15\linewidth}@{}>{\centering\arraybackslash}p{0.5\linewidth}@{}>{\centering\arraybackslash}p{0.1\linewidth}@{}}
    \toprule
     & {\bf Layout w/ DivCon}& {\bf Image w/ DivCon}& 
        \begin{tabular}{@{}c@{}} {\bf Numerical}\\ \hline 
        \begin{tabular}{@{}>{\raggedleft\arraybackslash}p{0.2\linewidth}@{}>{\raggedleft\arraybackslash}p{0.2\linewidth}@{}>{\raggedleft\arraybackslash}p{0.2\linewidth}@{}>{\raggedleft\arraybackslash}p{0.2\linewidth}@{}} Precision & Recall & F1 & Acc. \end{tabular}\end{tabular} & 
        \begin{tabular}{@{}r@{}} {\bf Spatial}\\ \hline Acc. \end{tabular}\\
    \midrule
      (1)& -& -& \begin{tabular}{@{}>{\raggedleft\arraybackslash}p{0.2\linewidth}@{}>{\raggedleft\arraybackslash}p{0.2\linewidth}@{}>{\raggedleft\arraybackslash}p{0.2\linewidth}@{}>{\raggedleft\arraybackslash}p{0.2\linewidth}@{}} 60.15& 55.87& 57.93& 15.96\end{tabular} &32.19\\
     (2)& \checkmark& -& \begin{tabular}{@{}>{\raggedleft\arraybackslash}p{0.2\linewidth}@{}>{\raggedleft\arraybackslash}p{0.2\linewidth}@{}>{\raggedleft\arraybackslash}p{0.2\linewidth}@{}>{\raggedleft\arraybackslash}p{0.2\linewidth}@{}} 76.16& 60.31& 69.96& 23.32\end{tabular} &47.06\\
     (3)& -& \checkmark& \begin{tabular}{@{}>{\raggedleft\arraybackslash}p{0.2\linewidth}@{}>{\raggedleft\arraybackslash}p{0.2\linewidth}@{}>{\raggedleft\arraybackslash}p{0.2\linewidth}@{}>{\raggedleft\arraybackslash}p{0.2\linewidth}@{}} 68.15 & 59.63 & 62.27 & 20.42 \end{tabular} &41.37\\
     (4)& \checkmark& \checkmark& \begin{tabular}{@{}>{\raggedleft\arraybackslash}p{0.2\linewidth}@{}>{\raggedleft\arraybackslash}p{0.2\linewidth}@{}>{\raggedleft\arraybackslash}p{0.2\linewidth}@{}>{\raggedleft\arraybackslash}p{0.2\linewidth}@{}} \textbf{78.65}& \textbf{63.41}& \textbf{70.22}& \textbf{29.17}\end{tabular} &\textbf{54.21}\\
    \bottomrule
  \end{tabular} }
  \caption{Ablation study of our DiVCon on HRS benchmark. ``Layout w/ DivCon": Layout prediction with divide-and-conquer strategy. ``Image w/ DivCon": Layout-to-Image generation with divide-and-conquer strategy.}
  \label{tab:ablation}
\end{table}

\subsection{Limitations and Discussions}

\begin{figure}[ht]
  \centering
  \includegraphics[width=0.7\linewidth]{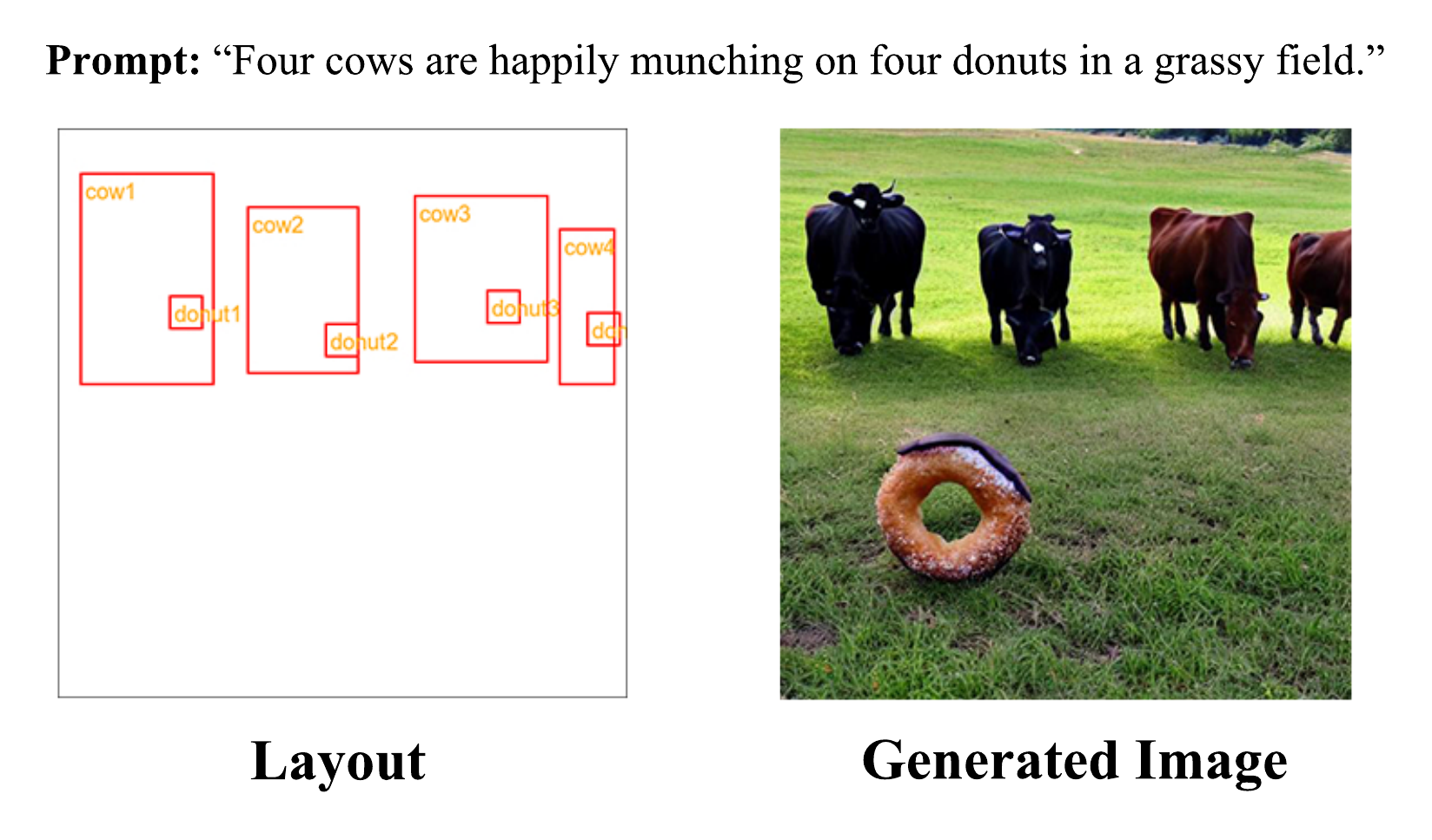}
  \vspace{-5pt}
  \caption{An illustration of failure case.}
  \label{fail example}
  \vspace{-0.2cm}
\end{figure}

As illustrated in Fig.~\ref{fail example}, when the text prompt describes objects with extreme overlap, the diffusion model may fail to generate all objects accurately. This limitation stems from the inherent capacity of layout-conditioned image generation models to handle overlapping boxes. Despite these challenges, our approach still demonstrates an improved quality in layout prediction as compared to the base model. 

\section{Conclusion}
In this work, we introduce DivCon, a novel divide-and-conquer approach that divides the text-to-image generation task into multiple subtasks to achieve higher image generation quality. Specifically, the layout prediction is divided into two steps to conduct numerical \& spatial reasoning and bounding box prediction, {which enables even lightweight LLMs to match the reasoning performance of large-scale models.} Layout-to-image generation is achieved via a progressive manner to synthesize objects in an easy-to-hard fashion. 
Comprehensive experiments demonstrate that our DivCon outperforms existing state-of-the-art grounded text-to-image models in terms of both image quality and prompt fidelity.

\bibliography{mybibfile}

\clearpage
\appendix

\twocolumn[{%
  \begin{center}
    \vspace*{1em}
    {\LARGE\bfseries Appendix}\\[0.5em]
  \end{center}
  \vspace{0.75em}
}]
In the appendix, we analyze the attention distribution during the second-round forward pass of the diffusion model, and provide example prompts for LLMs.

\section{Attention Distribution in Second-round Generation}
\begin{figure}[h]
  \centering
  \includegraphics[width=1\linewidth]{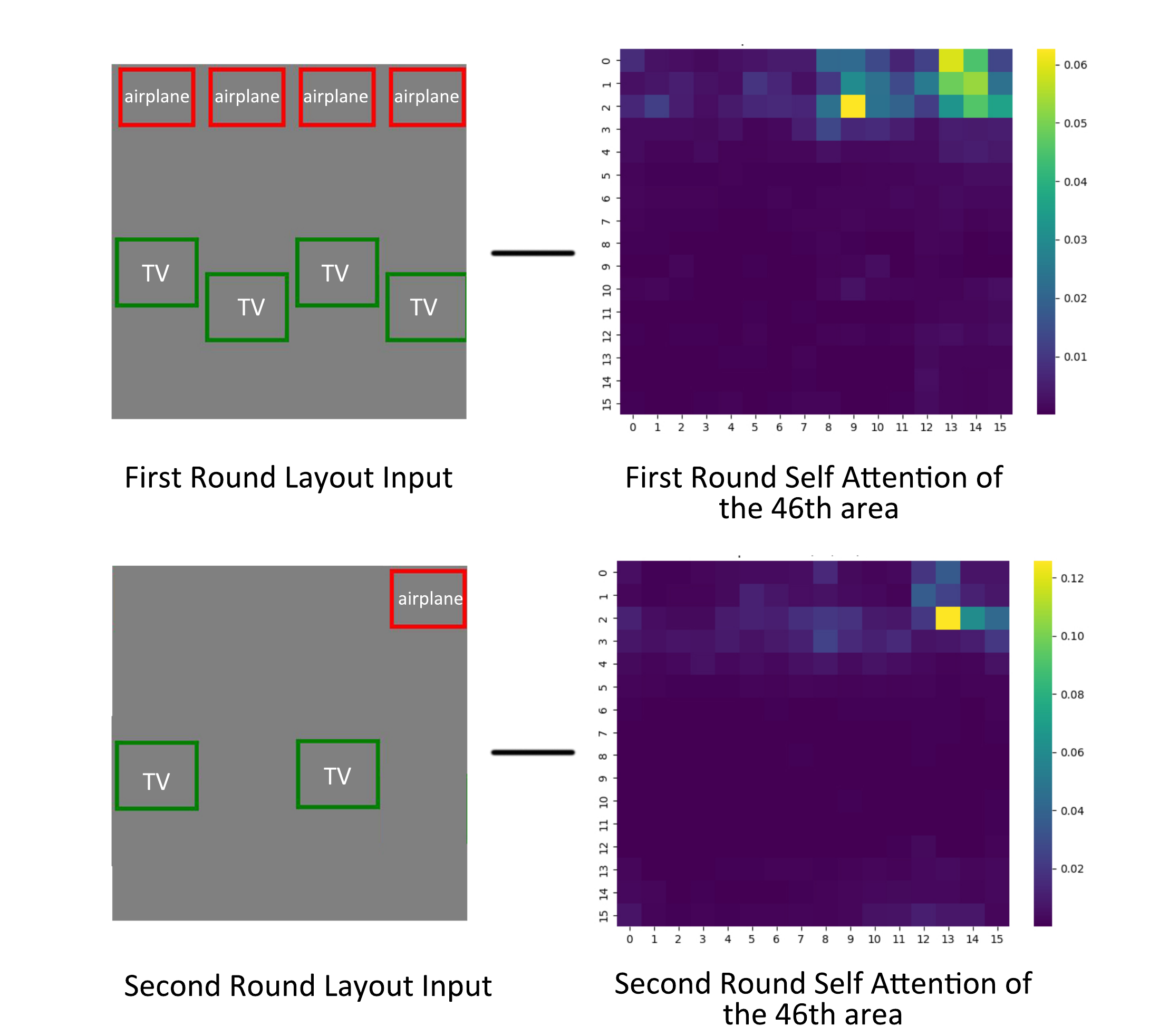}
  \caption{The attention map obtained in the first and second round generation.
  }
  \label{self_att}
\end{figure}

\textbf{Gated Self-Attention.} Our model leverages a pretrained layout-to-image generation model (GLIGEN \cite{li2023gligen}), which utilizes a gated self-attention mechanism to include layout information as new conditions. The attention is performed over the concatenation of visual tokens $v$ and grounded layout tokens $h$. 
Since only bad samples are considered during the second-round generation to calculate the attention scores, the model can focus more on these regions. To further validate this point, we visualize the attention maps in the last time step. As illustrated in Fig.~\ref{self_att}, the attention map obtained in the first round contains high scores outside the groundtruth box. In contrast, the attention scores are concentrated in the corresponding box, which ultimately improves the quality of these hard regions.

\begin{figure}[h]
  \centering
  \includegraphics[width=1\linewidth]{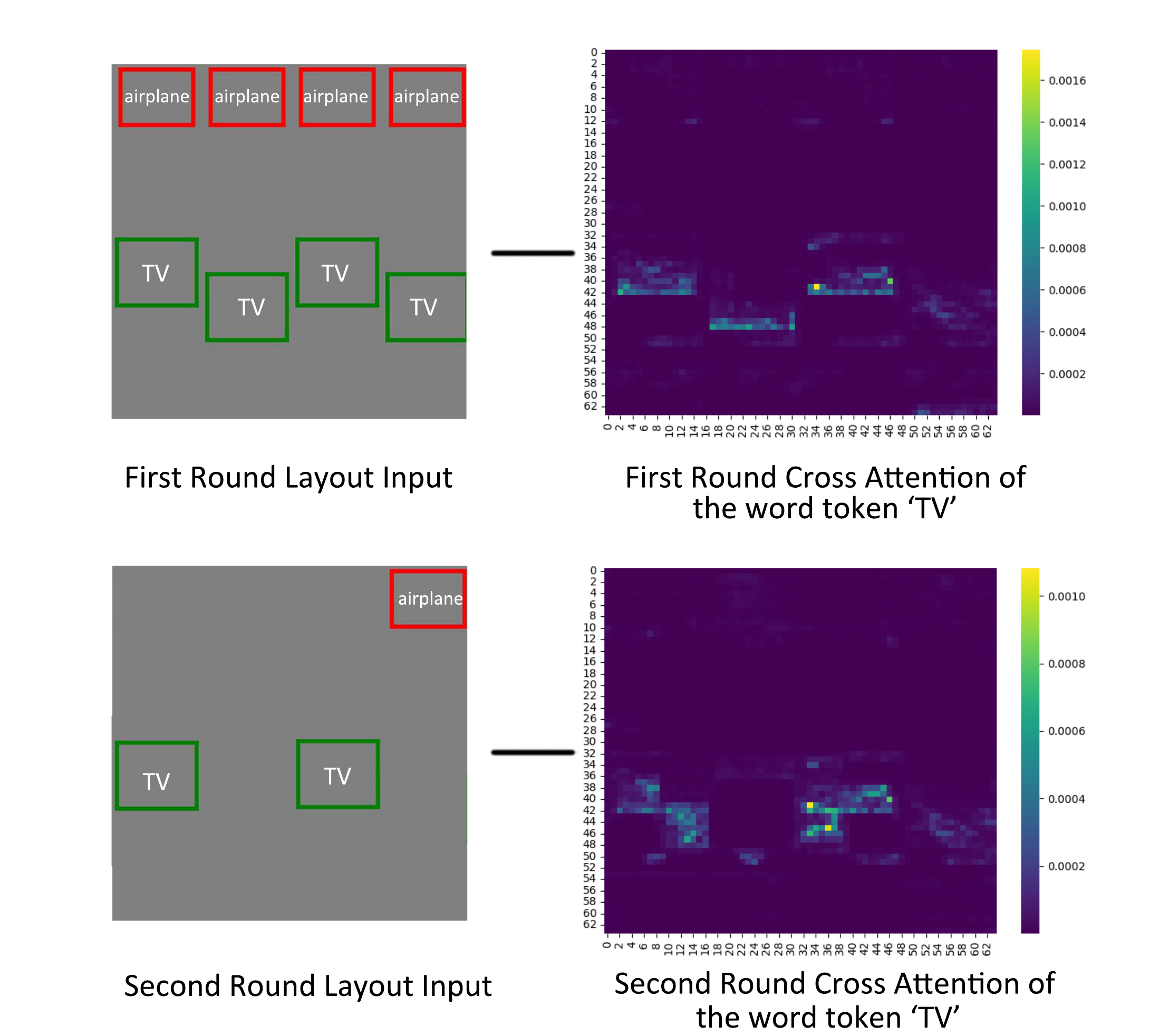}
  \caption{Attention maps for word token `TV' in the first and second round generation.
  }
  \label{cross_att}
\end{figure}

\begin{table}[h]
    \renewcommand{\arraystretch}{1.2} 
    \centering
    \begin{tabular}{ccccccc}\hline
         Number of & \multicolumn{4}{c}{Numerical} & \multicolumn{2}{c}{Spatial} \\ 
        Exemplars &  Pre.&  Rec.&  F1.&  Acc.& \multicolumn{2}{c}{Acc.}\\  \hline
         1 & 93.25 & 89.71 & 91.45 & 79.14 & 90.15 &  \\
         2 & 94.17 & 89.63 & 91.80 & 80.54 &  91.87 \\
         3 & 95.20 & 87.19 & 91.02 & 81.21 & 91.90 &  \\
         4 & 94.56 & 88.72 & 91.55 & 79.21 &  89.77 \\ \hline
    \end{tabular}
    \caption{Ablation of the number of exemplars on the HRS dataset.}
    \label{tab:Exemplars_ablation}
\end{table}

\begin{table}[h]
    \renewcommand{\arraystretch}{1.2} 
    \centering
    \begin{tabular}{ccccccc}\hline
         w/ Negation & \multicolumn{4}{c}{Numerical} & \multicolumn{2}{c}{Spatial} \\ 
        Exemplar &  Pre.&  Rec.&  F1.&  Acc.& \multicolumn{2}{c}{Acc.}\\ \hline
         \checkmark & 94.17 & 89.63 & 91.80 & 80.73 &  91.87 \\
                  - & 92.15 & 86.79 & 89.37 & 77.24 &  86.35 \\ \hline
    \end{tabular}
    \caption{Ablation of the negation exemplar on the HRS dataset.}
    \label{tab:negation_ablation}
\end{table}

\textbf{Cross-Attention Refocusing (CAR) Loss.} Following \cite{attention-refocusing}, our model utilizes a CAR loss for optimization. The CAR loss is formulated to enhance the attention scores within the designated box regions while discouraging them outside these areas. It is computed individually for each input box and subsequently aggregated across all boxes. In our second-round generation, CAR loss is selectively applied to the boxes of hard samples, thereby increasing the relative impact of this loss and improving generation quality in these areas. As shown in Fig.~\ref{cross_att}, the word token `TV' pays more attention to the area of hard samples instead of being evenly distributed across all samples.

\section{Layout Generation Prompting}
\begin{figure}[h]
  \centering
  \includegraphics[width=1\linewidth]{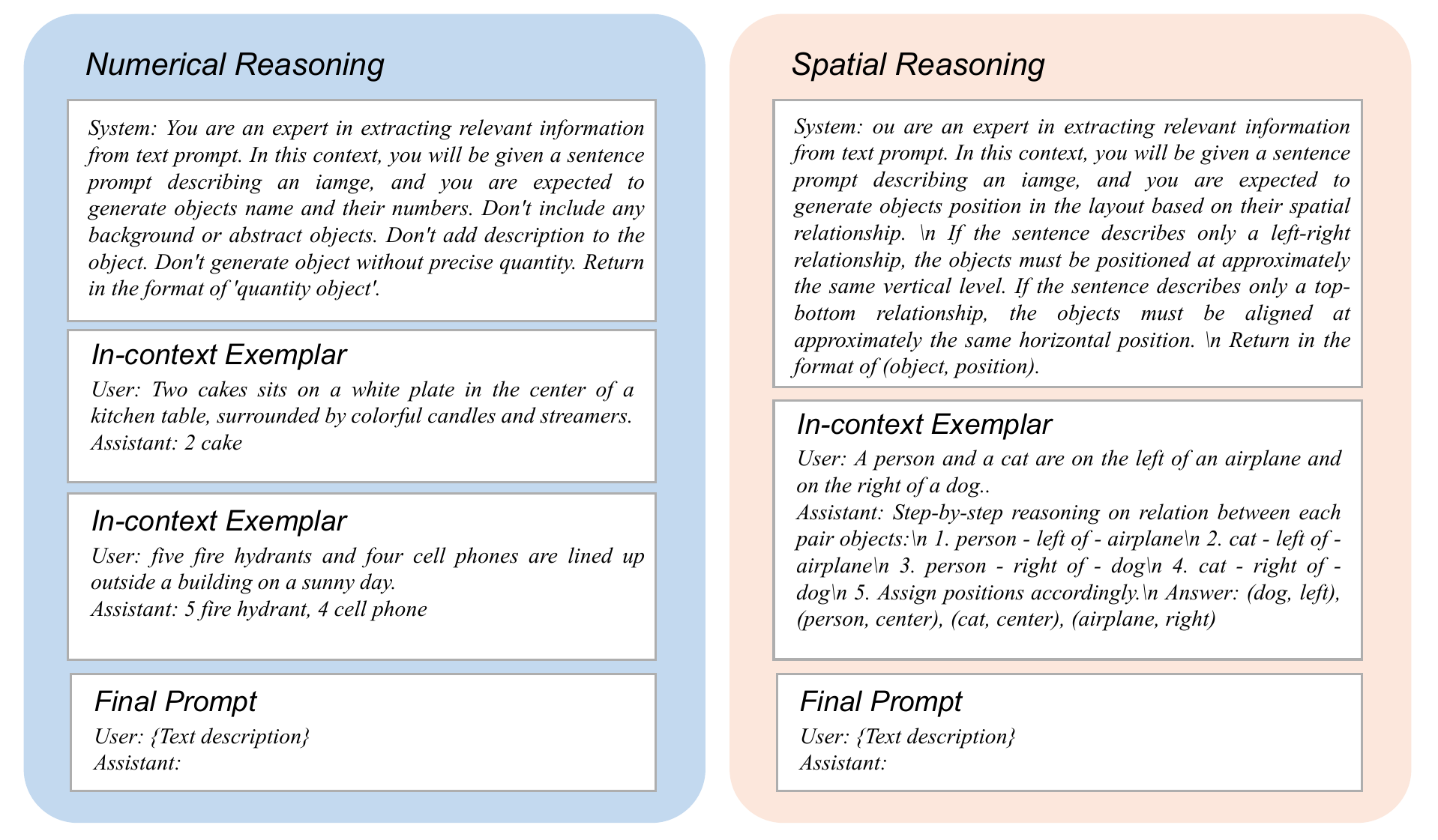}
  \caption{The example prompts for numerical and spatial reasoning.
  }
  \label{reasoning}
\end{figure}

\begin{figure}[h]
  \centering
  \includegraphics[width=1\linewidth]{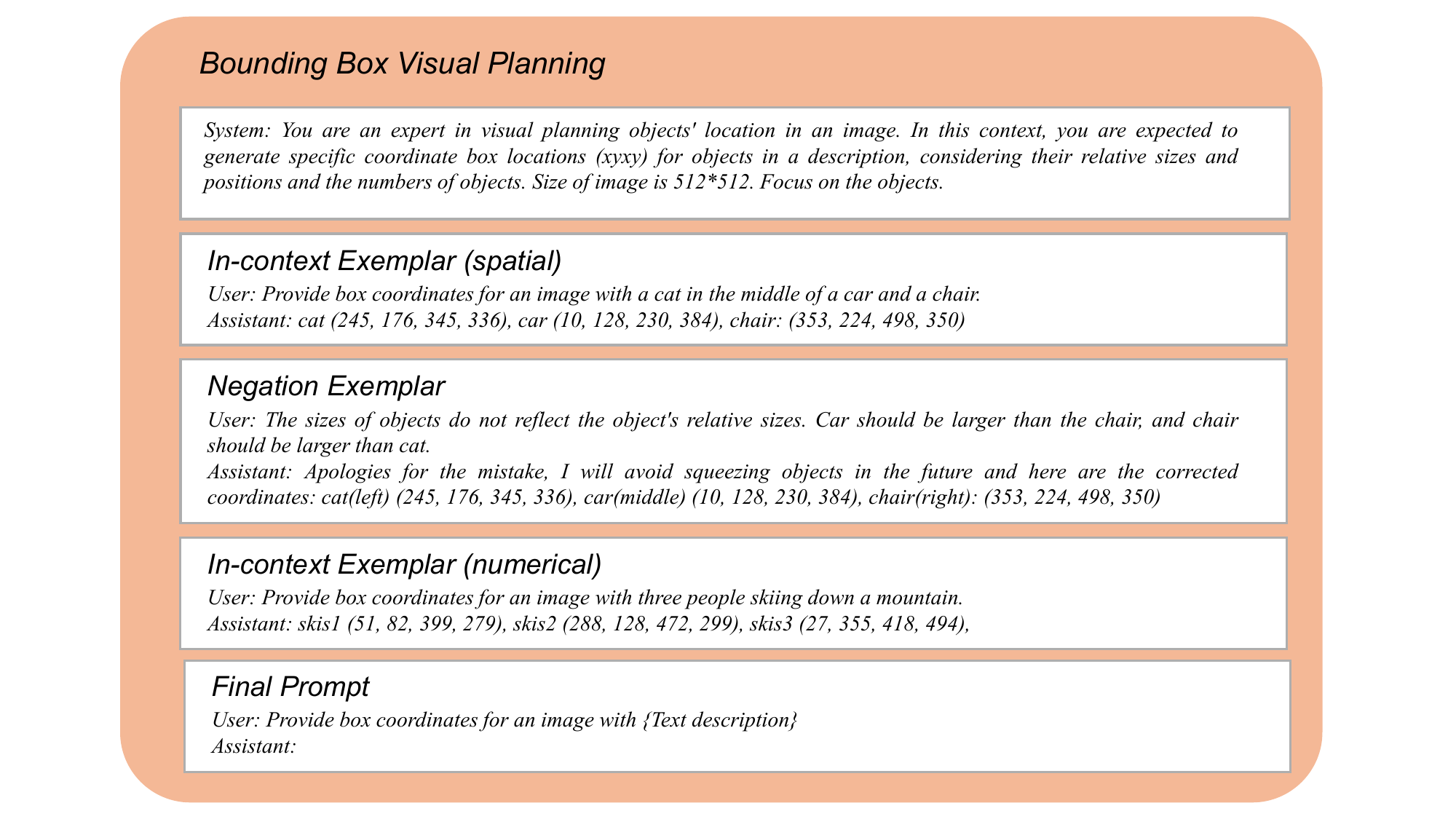}
  \caption{The example prompts for bounding box visual planning.
  }
  \label{visual planning}
\end{figure}

We design prompts for numerical and spatial reasoning respectively, including an in-context exemplar and asks LLM to extract numerical information (e.g. 5 fire hydrant, 4 cell phone) or location information (e.g. dog, left; person, center; cat, center; airplane, right;) of objects in the image (Fig. \ref{reasoning}). The bounding box visual planning shares one prompt including 2 in-context exemplars and one negation exemplar which assist the LLM to consider the relative sizes of objects (Fig. \ref{visual planning}).


On the HRS benchmark, we conduct ablation experiments of the negation exemplar and the number of in-context exemplars. Table \ref{tab:Exemplars_ablation} compares layout accuracy using varying numbers of in-context exemplars. The results show the number of exemplars has minimal impact on the layout accuracy. Consequently, we adopt two exemplars in our experiments. Table \ref{tab:negation_ablation} shows that incorporating a negation exemplar yields a 3\% improvement in terms of numerical accuracy and a 4\% gain on spatial accuracy. Consequently, we included the negation exemplar as the default prompt design.

\end{document}